\documentclass[letterpaper, 10 pt, conference]{IEEEtran} 
\usepackage[utf8]{inputenc}
\pdfoutput=1
\IEEEoverridecommandlockouts

\usepackage{amsmath}
\usepackage{amssymb}
\usepackage{bm}
\usepackage{color}
\usepackage{xcolor}
\usepackage{graphicx}
\usepackage{siunitx}
\usepackage{url}
\usepackage{balance}
\usepackage[ruled]{algorithm2e}
\usepackage{subfigure}
\usepackage{multirow}
\usepackage{tabularx}

\newcommand*{\rom}[1]{(\expandafter{\romannumeral #1\relax})}
\newcolumntype{Y}{>{\centering\arraybackslash}X}

\definecolor{somegray}{rgb}{0.5, 0.5, 0.5}
\newcommand{\darkgrayed}[1]{\textcolor{somegray}{#1}}
\makeatletter
\newcommand*\titleheader[1]{\gdef\@titleheader{#1}}
\AtBeginDocument{%
  \let\st@red@title\@title
  \def\@title{%
    \vskip-1.5em
    \bgroup\normalfont\large\centering\@titleheader\par\egroup
    \vskip-0.5em\st@red@title}
}
\makeatother

\titleheader{\darkgrayed{This paper has been accepted for publication at the\\
IEEE/RSJ International Conference on Intelligent Robots and Systems (IROS), 2020.
\copyright IEEE}}

\title{\vspace{18pt}Faster than FAST:\\GPU-Accelerated Frontend for High-Speed VIO\vspace{-6pt}}
\author{Bal\'{a}zs Nagy, Philipp Foehn, Davide Scaramuzza \thanks{All authors are with
the Robotics and Perception Group, Dep. of Informatics, University of
Zurich, and Dep. of Neuroinformatics, University of Zurich and ETH Zurich,
Switzerland - http://rpg.ifi.uzh.ch.
Their work was supported
by the SNSF-ERC Starting Grant and the Swiss National Science Foundation through the National Center of Competence in Research (NCCR)
Robotics.}\vspace{-12pt}}
\date{January 2020}

\begin{document}

\maketitle

\begin{abstract}
The recent introduction of powerful embedded graphics processing units (GPUs) has allowed for unforeseen improvements in real-time computer vision applications. It has enabled algorithms to run onboard, well above the standard video rates, yielding not only higher information processing capability, but also reduced latency.
This work focuses on the applicability of efficient low-level, GPU hardware-specific instructions to improve on existing computer vision algorithms in the field of visual-inertial odometry (VIO).
While most steps of a VIO pipeline work on visual features, they rely on image data for detection and tracking, of which both steps are well suited for parallelization.
Especially non-maxima suppression and the subsequent feature selection are prominent contributors to the overall image processing latency.
Our work first revisits the problem of non-maxima suppression for feature detection specifically on GPUs, and proposes a solution that selects local response maxima, imposes spatial feature distribution, and extracts features simultaneously.
Our second contribution introduces an enhanced FAST feature detector that applies the aforementioned non-maxima suppression method.
Finally, we compare our method to other state-of-the-art CPU and GPU implementations, where we always outperform all of them in feature tracking and detection, resulting in over 1000fps throughput on an embedded Jetson TX2 platform.
Additionally, we demonstrate our work integrated into a VIO pipeline achieving a metric state estimation at $\bm{\sim}$200fps.
\end{abstract}

\urlstyle{same}
\vspace{3pt}
Code available at: \url{https://github.com/uzh-rpg/vilib}

\section{Introduction}
\vspace{-8pt}
\subsection{Motivation}
As technology became increasingly affordable, vision-based motion tracking has proven its capabilities not only in robotics applications, such as autonomous cars and drones but also in virtual (VR) and augmented (AR) reality and mobile devices.
While visual-inertial odometry (VIO) prevails with its low cost, universal applicability, and increasing maturity and robustness, it is still computationally expensive and introduces significant latency.
This latency impacts e.g. VR/AR applications by introducing motion sickness, or robotic systems by constraining their control performance.
The latter is especially true for aerial vehicles with size and weight constraints limiting the available computation power while requiring real-time execution of the VIO and control pipeline to guarantee stable, robust, and safe operation.
Besides latency, one may also witness a disconnect between the available sensor capabilities (both visual and inertial) and the actual information processing capabilities of mobile systems.
While off-the-shelf cameras are capable of capturing images above 100fps, many algorithms and implementations are not able to handle visual information at this rate.
By lowering the frame processing times, we can simultaneously minimize latency and also reduce the neglected visual-inertial information.

In particular, embedded systems on drones or AR/VR solutions cannot rely on offline computations and therefore need to use all their available resources efficiently.
Various heterogeneous embedded solutions were introduced, offering a range of computing architectures for better efficiency.
There are three popular heterogeneous architectures:
\rom{1} the first one uses a central processing unit (CPU) with a digital signal processor (DSP), and therefore it is restricted in its set of tasks;
\rom{2} the second one combines a CPU with programmable logic (e.g. FPGA), which is versatile but increases development time;
\rom{3} the third solution is the combination of a CPU with a GPU, which is not only cost-efficient but also excels in image processing tasks since GPUs are built for highly parallel tasks.

On these grounds, our work investigates feature detection and tracking on CPU-GPU platforms, where we build up the image processing from the GPU hardware's perspective.
We present a feasible work-sharing between the CPU and a GPU to achieve significantly lower overall processing times.

\begin{figure}[!t]
\centering
\includegraphics[width=0.47\textwidth, trim=0 20 0 0, clip]{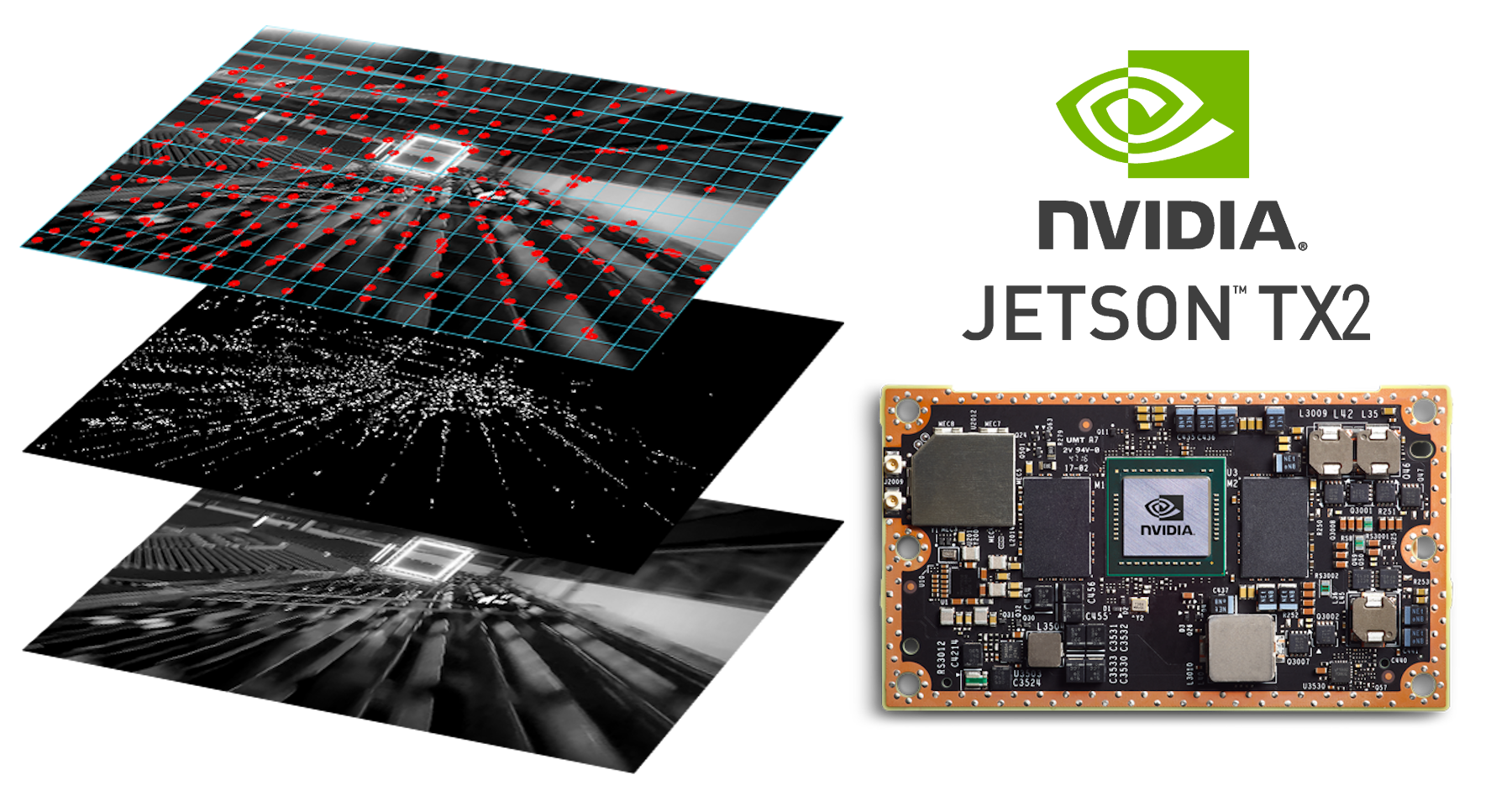}
\caption{Our method introduces a novel non-maxima suppression scheme exploiting GPU parallelism and low-level instructions, applied for GPU-optimized feature detection and tracking.
We demonstrate a feature detection and tracking rate of \textbf{over 1000fps} on an embedded Jetson TX2 platform.}
\vspace{-10pt}
\end{figure}

\subsection{Related Work}
We recapitulate previous approaches according to the building blocks of a VIO pipeline front-end: feature detection, non-maxima suppression, and feature tracking.

\subsubsection{Feature Detection}
Over the years, feature detection has not changed significantly.
Pipelines commonly use Harris \cite{Harris88acombined}, Shi-Tomasi \cite{Shi:1993:GFT:866676}, or FAST features \cite{rosten_2006_machine} with one of three FAST corner scores.
Harris and Shi-Tomasi are less sensitive to edges and are also widely used independently as corner detectors.
They both share the same principles, but their metric of cornerness differs.
ORB \cite{Rublee:2011:OEA:2355573.2356268}, as an extension to FAST has also appeared in VIO pipelines, presenting a reasonable, real-time alternative to SIFT \cite{Lowe:2004:DIF:993451.996342} and SURF \cite{Bay:2008:SRF:1370312.1370556}.
Amongst the above, undoubtedly FAST presents the fastest feature detector.
Two variations were proposed from the original authors in the form of FAST \cite{rosten_2006_machine} and FAST-ER \cite{rosten_2008_faster}, where the latter greatly improves on repeatability while still maintaining computational efficiency.
To the best of our knowledge, the fastest CPU implementation of the FAST detector is KFAST \cite{Komrad2006github}, which showcases more than 5$\times$ speedup over the original implementation.
On the GPU, we are aware of two optimized implementations within OpenCV \cite{opencv_library} and ArrayFire \cite{Yalamanchili2015}.
Both employ lookup tables to speed up the decision process for determining a point's validity: the latter uses a 64-kilobyte lookup table, while the former an 8-kilobyte one.
Although both solutions provide fast cornerness decision, neither of them guarantees spatial feature distribution as they both stop extracting features once the feature count limit is reached.

\subsubsection{Non-Maxima Suppression}
Non-maxima suppression can be considered a local maximum search within each candidate's Moore neighborhood.
The Moore neighborhood of each pixel-response is its square-shaped surrounding with a side length of $(2n+1)$.
Suppression of interest point candidates has been studied extensively in \cite{1699659, Pham10}, and recently, they have also been reviewed in machine learning applications \cite{7471831}.
The complexity of the proposed algorithms is determined based on the number of comparisons required per interest point.
This algorithm requires $(2n+1)^2$ comparisons, as the comparisons follow the raster scan order.
Förstner and Gülch \cite{forstner1987fast} proposed performing the comparisons in spiral order.
Theoretically, the number of comparisons does not change, however, the actual number of comparisons plummeted, because most candidates can be suppressed in a smaller 3x3 neighborhood first, and only a few points remain.
Neubeck \cite{1699659} proposed several algorithms to push down the number of comparisons to almost 1 in the worst-case.
Pham \cite{Pham10} proposed another two algorithms (scan-line, and quarter-block partitioning) that drove down the number of comparisons below 2, not only for larger ($n\ge5$) but also for small neighborhood sizes ($n<5$).
All these approaches first try to perform reduction to a local maximum candidate in a smaller neighborhood, then perform neighborhood verification for only the selected candidates.
However, they do not ensure any (spatial) feature distribution and possibly output a large set of feature candidates.

\subsubsection{Feature tracking}
Feature tracking may be divided into three different categories: \rom{1} feature matching, \rom{2} filter-based tracking, and \rom{3} differential tracking.
Feature matching \rom{1} applies feature extraction on each frame followed by feature matching, which entails a significant overhead.
Moreover, the repeatability of the feature detector may adversely influence its robustness.
However, there are well-known pipelines, opting for this approach \cite{4209642, Leutenegger:2015:KVO:2744155.2744163, journals/corr/Mur-ArtalMT15}.
\cite{7353389} tracks the features using filters \rom{2}, which contain the feature location in its state (e.g. via bearing vectors), and follows features with consecutive prediction and update steps.
The third differential \rom{3} approaches aim to directly use the pixel intensities and minimize a variation of the photometric error.
From the latter kind, the Lucas-Kanade tracker \cite{Lucas81ijcai, Baker-2002-8493, Baker-2003-8809} became ubiquitous in VIO pipelines \cite{6906584, Forster:2017:SSV:3083770.3083863} due to its efficiency and robustness.
As it directly operates on pixel intensity patches, GPU adaptations appeared early on.
\cite{4563089} implements a translational displacement model on the GPU with intensity-gain estimation.
\cite{5457608, 5354093} go even further: they propose an affine-photometric model coupled with an inertial measurement unit (IMU) initialization scheme: the displacement model follows an affine transformation, in which the parameters are affected by the IMU measurements between consecutive frames, while the pixel intensities may also undergo an affine transformation.

\subsection{Contributions}
Our work introduces a novel non-maxima suppression building upon \cite{forstner1987fast} but also exploiting low-level GPU instruction primitives, completed by a GPU-optimized implementation of the FAST detector with multiple scores.
Our method combines the feature detection and non-maxima suppression, guaranteeing uniform feature distribution over the whole image, which other approaches need to perform in an extra step.
Additionally, we combine our frontend with a state-of-the-art VIO bundle-adjustment backend.
All contributions are verified and thoroughly evaluated using the EuRoC dataset \cite{Burri25012016} on the Jetson TX2 embedded platform and a laptop GPU.
Throughput capabilities of over 1000fps are demonstrated for feature detection and tracking, and $\sim$200fps for a VIO pipeline recovering a metric state estimate.

\section{Methodology}

\subsection{Preliminaries on parallelization}
We briefly introduce the fundamentals of the Compute Unified Device Architecture, or shortly CUDA, which is a parallel computing platform and programming model proprietary to NVIDIA.
CUDA allows developers to offload the central processing unit, and propagate tasks to the GPU even for non-image related computations. Latter is commonly referred to as general-purpose GPU (GPGPU) programming.

The NVIDIA GPU architecture is built around a scalable array of multithreaded streaming multiprocessors (SMs) \cite{cudacprogrammingguide}. Each SM has numerous streaming processors (SP), that are lately also called CUDA cores. GPUs usually have 1-20 streaming multiprocessors and 128-256 streaming processors per SM. In addition to the processing cores, there are various types of memories available (ordered by proximity to the processing cores): register file, shared memory, various caches, off-chip device, and host memory.

NVIDIA’s GPGPU execution model introduces a hierarchy of computing units: threads, warps, thread blocks, and thread grids. The smallest unit of execution is a thread. Threads are grouped into warps: each warp consists of 32 threads.
Warps are further grouped into thread blocks. One thread block is guaranteed to be executed on the same SM. Lastly, on top of the execution model is the thread grid. A thread grid is an array of thread blocks. Thread blocks within a thread grid are executed independently from each other.

The instruction execution on the GPU needs to be emphasized: every thread in a warp executes the same instruction in a lock-step basis. NVIDIA calls this execution model Single Instruction Multiple Threads (SIMT). It also entails, that \textit{if/else} divergence within a warp causes serialized execution.

The underlying GPU hardware occasionally undergoes significant revisions, hence the differences between GPUs need to be tracked. NVIDIA introduced the notion of Compute Capability accompanied by a codename to denote these differences.
With the introduction of the NVIDIA Kepler GPU microarchitecture, threads within the same warp can read from each other's registers with specific instructions.
Our work focuses on these warp-level primitives, more specifically, highly-efficient communication patterns for sharing data between threads in the same warp.
In previous GPU generations, threads needed to turn to a slower common memory (usually the shared memory) for data sharing, which resulted in significant execution latencies.
However, with the introduction of the Kepler architecture it became possible to perform communication within the warp first, and only use the slower memory on higher abstractions of execution, i.e. within thread blocks and then within the thread grid.
In Table \ref{table:gpu_comm} we summarized the available fastest memories for exchanging data between execution blocks on a GPU.

\begin{table}[t]
\setlength{\tabcolsep}{3pt}
\centering
\caption{Memory selection for the fastest communication}
\label{table:gpu_comm}
\footnotesize
\begin{tabularx}{\linewidth}{l|Y|Y}
\hline
\textbf{Execution Unit} & \textbf{Execution Unit} & \textbf{Fastest Memory} \\
\hline
\hline
Threads within warp & identical SM & registers \\
\hline
Warps within thread block & identical SM & shared memory \\
\hline
Thread blocks & any SM & global memory \\
\hline
\end{tabularx}
\end{table}

\subsection{Feature detector overview}
GPUs are particularly well-suited for feature detection, which can be considered a stencil operation amongst the parallel communication patterns.
In a stencil operation, each computational unit accesses an input element (e.g. pixel) and its close neighborhood in parallel.
Therefore, the image can be efficiently divided amongst the available CUDA cores, such that the memory accesses are coalesced, leading to highly effective parallelization.
For feature detection, the input image is first subsampled to acquire an image pyramid.
Then, for each image resolution, two functions are usually evaluated at each pixel: a coarse corner response function (CCRF), and a corner response function (CRF).
CCRF serves as a fast evaluation that can swiftly exclude the majority of candidates so that a slower CRF function only receives candidates that passed the first verification.
Once every pixel has been evaluated within the ROI, non-maxima suppression is applied to select only the local maxima.
We summarized the general execution scheme of feature detector algorithms in Algorithm \ref{alg:feature_det}.
\begin{algorithm}[t]
\small
\For{Every scale}{
  \For{Every pixel within the region of interest (ROI)}{
    \If{Coarse Corner Response Function (CCRF)}{
     Corner Response Function (CRF)
    }
  }
  Non-max. suppression within neighborhood (NMS)
}
\textcolor{orange}{\textbullet} Non-max. suppression within cell (NMS-C)
\caption{Generalized Feature Detection}
\label{alg:feature_det}
\end{algorithm}
It was discovered in \cite{1315094, Scaramuzza:2009:RMV:1703435.1703515, 6153423}, that uniform feature distribution on image frames improves the stability of VIO pipelines.
To fulfill this requirement, \cite{6906584} and \cite{Forster:2017:SSV:3083770.3083863} introduced the notion of 2D grid cells: the image is divided into rectangles with a fixed width and height.
Within each cell, there is only one feature selected - the feature whose CRF score is the highest within the cell.
Not only does this method distribute the features evenly on the image, but it also imposes an upper limit on the extracted feature count.
This is shown in Algorithm \ref{alg:feature_det}, including the augmentation~\textcolor{orange}{\textbullet}.

\subsection{Non-maxima suppression with CUDA}
\label{sec:met_nmscuda}
Feature selection within a cell can be understood as a reduction operation, where only the feature with the maximum score is selected.
Moreover, non-maxima suppression within the neighborhood can also be considered a reduction operation, where one reduces the corner response to a single-pixel location within finite neighborhoods.

Our approach divides the corner response map into a regular cell grid.
Within the grid on the first pyramid level, cells use a width of an integer multiple of 32, i.e. $32w$, because, on NVIDIA GPU hardware, a warp consists of 32 threads.
One line of a cell is referred to as a cell line, which can be split up into cell line segments with 32 elements.
We restrict the height of the cells to be $2^{l-1}h$, where $l$ is the number of pyramid levels utilized during feature detection.
By selecting an appropriate grid configuration $l$, $w$, and $h$, one can determine the maximum number of features extracted, while maintaining spatial distribution.

Within one cell line segment, one thread is assigned to process one pixel-response, i.e. one warp processes one entire cell line segment (32 threads for 32 pixel-responses).
While one warp can process multiple lines, multiple warps within a thread block cooperatively process consecutive lines in a cell.
As the corner response map is stored using a single-precision floating-point format in a pitched memory layout, the horizontal cell boundaries perfectly coincide with the L1-cache line boundaries, which maximizes the memory bus utilization whenever fetching complete cell line segments.

\begin{figure}[t]
  \centering
  \includegraphics[width=0.65\linewidth, keepaspectratio]{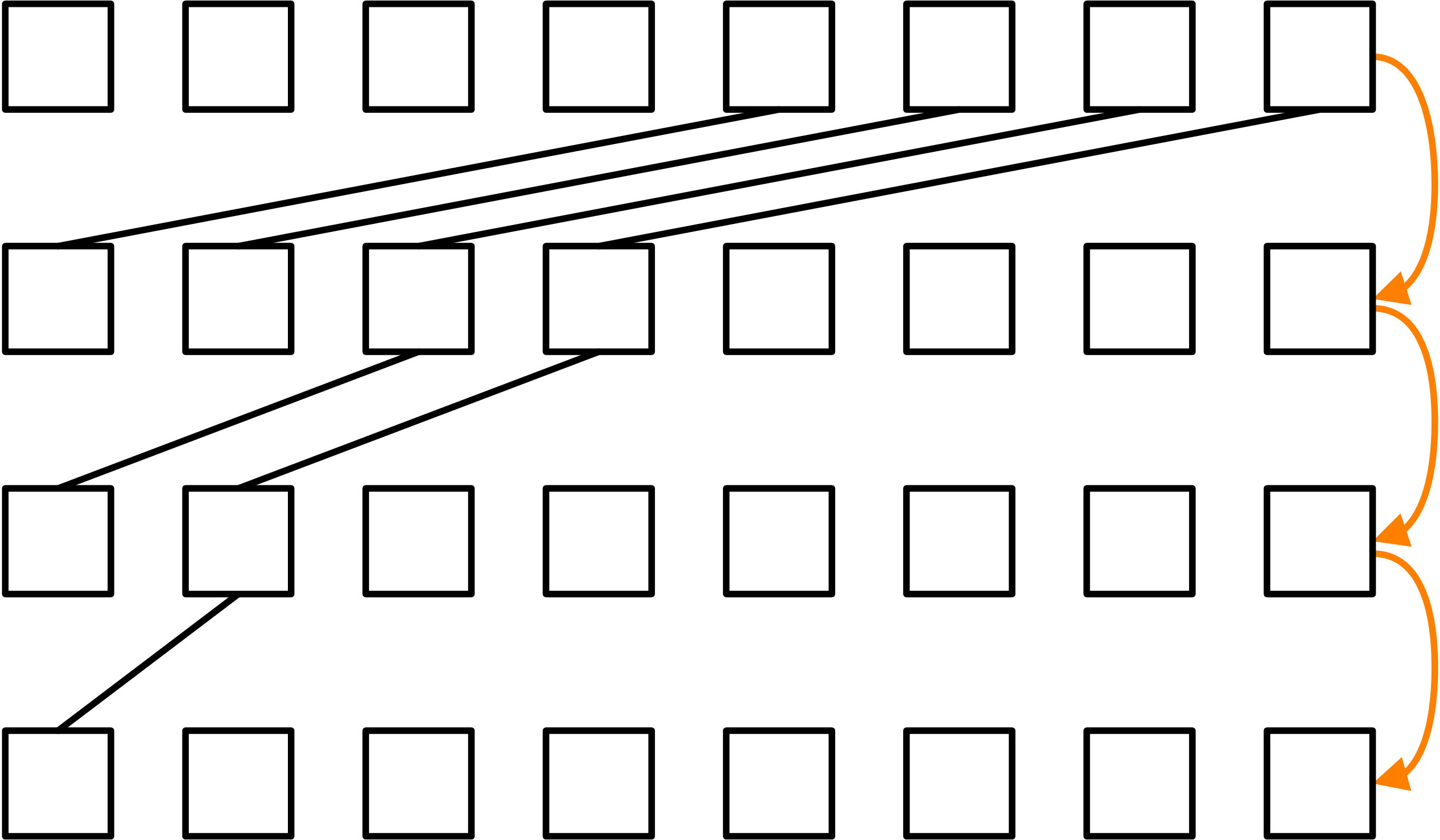}
  \caption{Warp-level communication pattern during cell maximum selection. At the end of the communication, thread 0 has the valid maximum.}
  \label{fig:shuffledownred}
\vspace{-5pt}
\end{figure}

For the simplicity of illustration, a 1:1 warp-to-cell mapping is used in a 32x32 cell.
As a warp reads out the first line of the cell, each thread within the warp, has acquired one pixel-response.
As the next operation, the entire warp starts the neighborhood suppression: the warp starts spiraling according to \cite{forstner1987fast}, and each thread verifies whether the response it has is the maximum within its Moore neighborhood.
Once the neighborhood verification finishes, a few threads might have suppressed their response.
However, no write takes place at this point, every thread stores its state (response score, and x-y location) in registers.
The warp continues with the next line, and repeats the previous operations: they read out the corresponding response, and start the neighborhood suppression, and update their maximum response if the new response is higher than the one in the previous line.
The warp continues this operation throughout all cell lines until it processed the entire cell.
Upon finishing, each thread has its maximum score with the corresponding 2D location.
However, as the 32 threads were processing individual columns, the maximum is only column-wise.
Therefore, the warp needs to perform a warp-level reduction to get the cell-wise maximum: they reduce the maximum score and location to the first thread (thread 0) using warp-level shuffle down reduction \cite{devblogfasterparallelreduction}.
The applied communication pattern is shown in Figure \ref{fig:shuffledownred}.
Thread 0 finally writes the result to global memory.

To speed up reduction, multiple (M) warps process one cell, therefore, after the warp-level reduction, the maximum is reduced in shared memory.
Once all warps within the thread block wrote their maximum results (score, x-y location) to their designated shared memory area, the first thread in the block selects the maximum for each cell and writes it to global memory, finishing the processing of this cell.

During pyramidical feature detection, we maintain only one grid.
On level 0 (original resolution), the above-specified algorithm applies.
On lower pyramid levels, we virtually scale the cell sizes, such that the applicable cell size on level k becomes $(\frac{32 \cdot w}{k},\frac{2^{l-1}h}{k})$.
In case the cell width falls below 32, one warp may process multiple lines: if the consecutive cell lines still belong to the same grid cell, the warp can analogously perform the warp-level reduction.
Since a lower pyramid level's resolution is half of its upper layer's resolution, we can efficiently recompute where a pixel response falls from lower pyramid levels on the original grid.
That is, when we identify a cell maximum on a lower pyramid level, the 2D position $(x,y)$ from the lower resolution can be scaled up to $(2^l x, 2^l y)$.

Looking back at Algorithm \ref{alg:feature_det}, our approach combines the regular neighborhood suppression (NMS) and cell maximum selection (NMS-C) into a single step.
It also differs from \cite{1699659, Pham10}, because we first perform candidate suppression within each thread in parallel, then reduce the remaining candidates amongst one cell.

\subsection{FAST feature detector}\label{sec:fast}
\begin{figure}[t]
  \centering
  \includegraphics[width=0.45\textwidth,trim={0 5 0 0},clip]{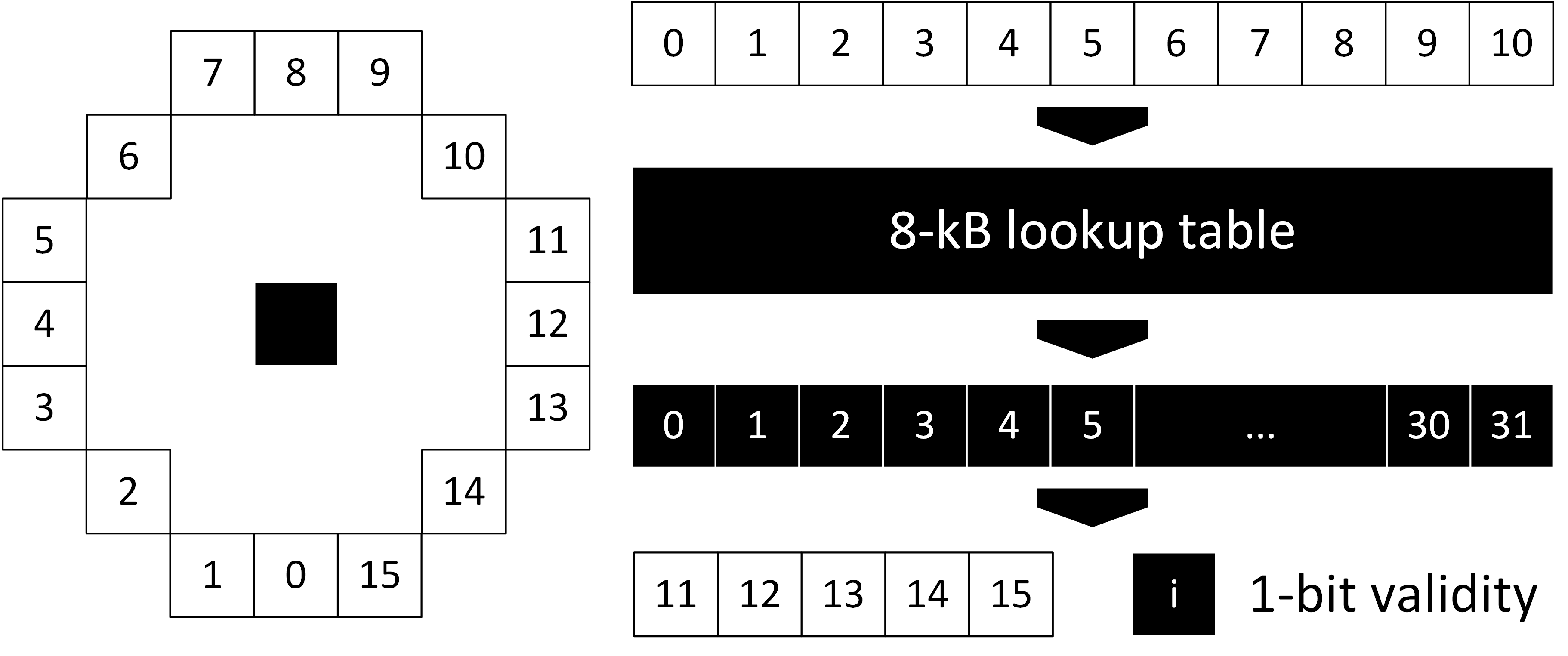}
  \caption{FAST corner point evaluation with an 8 kilobyte lookup table}
  \label{fig:fast}
  \vspace{-10pt}
\end{figure}

The FAST feature detector's underlying idea is simple: for every pixel location in the original image (excluding a minimum border of 3 pixels on all sides) we perform a segment test, in which we compare pixel intensities on a Bresenham circle with a radius of 3.
This Bresenham circle gives us 16 pixel-locations around each point (see Figure \ref{fig:fast}).
We give labels $L_x$ to these points based on a comparison between the center's and the actual point's intensity.

Given there is a continuous arc of at least N pixels that are labeled either \textit{darker} or \textit{brighter}, the center is considered a corner point.
To add more robustness to the comparisons, a threshold value ($\epsilon$) is also applied.
The comparisons are summarized in (\ref{eq:fast_comparison}).
Both the number of continuous pixels (N) and the threshold value ($\epsilon$) are tuning parameters.
\begin{equation}
L_x = \begin{cases}
\text{darker} & I_x < I_{center} - \epsilon\\
\text{similar} & I_{center} - \epsilon \leq I_x \leq I_{center} + \epsilon\\
\text{brighter} & I_{center} + \epsilon < I_x
\end{cases}
\label{eq:fast_comparison}
\end{equation}
\textbf{Avoiding Execution Divergence in FAST Calculation}

If each thread performed the comparisons from (\ref{eq:fast_comparison}) in NVIDIA's single-instruction-multiple-threads (SIMT) execution model, the comparison decisions in \textit{if/else}-instructions will execute different code blocks.
Since all threads execute the same instruction in a warp, some threads will be inactive during the \textit{if}-branch and others during the \textit{else}-branch.
This is called code divergence and reduces the throughput in parallelization significantly, but it can be resolved with a completely different approach: a lookup table (Figure \ref{fig:fast}).

Our approach stores the result of the 16 comparisons as a bit array, which serves as an index for the lookup table.
All possible 16-bit vectors are precalculated: a bit $b_x$ is '1' if the pixel intensity on the Bresenham circle $I_x$ is darker/brighter than the center pixel intensity $I_{center}$, and '0' if the pixel intensities are similar.
As the result is binary for all $2^{16}$ vectors, the answers can be stored on $2^{16}$ bits, i.e. 8 kilobytes.
These answers can be stored by using 4-byte integers, each of which store 32 combinations ($2^5$): 11 bits are used to acquire the address of the integer, and the 5 unused bits then select one bit out of the 32.
If the resulting bit is set, we proceed with the calculation of the corner score.

The literature distinguishes three different types of scores for a corner point: sum of absolute differences on the entire Bresenham circle (SAD-B); sum of absolute differences on the continuous arc (SAD-A) \cite{rosten_2006_machine}; maximum threshold ($\epsilon$) for which the point is still considered a corner point (MT) \cite{rosten_2008_faster}.
The corner score is 0 if the segment test fails.

Our approach compresses the validity of each 16-bit combination into a single bit, resulting in an 8-kilobyte lookup table, for which the cache-hit ratio is higher than the work presented in \cite{Yalamanchili2015}.
This results from the increased reuse of each cache line that is moved into the L1 (and L2) caches, therefore improving the access latency.

\subsection{Lucas-Kanade Feature Tracker}
Our approach deploys the pyramidical approximated simultaneous inverse compositional Lucas-Kanade algorithm as feature tracker.
The Lucas-Kanade \cite{Lucas81ijcai} algorithm minimizes the photometric error between a rectangular patch on a template and a new image by applying a warping function on the image coordinates of the new image.
The inverse compositional algorithm is an extension that improves on the computational complexity per iteration \cite{Baker-2002-8493} by allowing to precompute the Hessian matrix and reuse it in every iteration.
The simultaneous inverse compositional Lucas-Kanade adds the estimation of affine illumination change.
However, as the Hessian becomes the function of the appearance estimates, it cannot be precomputed anymore, which makes this approach even slower than the original Lucas-Kanade.
Therefore, our approach applies the approximated version, where the appearance parameters are assumed to not change significantly, and hence the Hessian can be precomputed with their initial estimates \cite{Baker-2003-8809}.

We use a translational displacement model $\bm{t}$  with affine intensity variation estimation $\bm{\lambda}$.
The complete set of parameters are $\bm{q} = [\bm{t}, \bm{\lambda}]^\intercal = [t_x, t_y, \alpha, \beta]^\intercal$, where $t_x, t_y$ are the translational offsets, while $\alpha, \beta$ are the affine illumination parameters, resulting in the warping
\begin{equation}
\bm{W}(\bm{x}, \bm{t}) = 
\begin{pmatrix}
x + t_x \\
y + t_y
\end{pmatrix}.
\label{eq:lk_translational}
\end{equation}
The per-feature photometric error that we try to minimize for each feature with respect to $\Delta \bm{q} = [ \Delta\bm{t}, \Delta\bm{\lambda} ]$ is
\begin{equation}
\begin{aligned}
\min \sum_{\bm{x} \in N} \Big[ T(\bm{W}( \bm{x}, \Delta \bm{t})) - I(\bm{W}(\bm{x},\bm{t})) + \\ (\alpha + \Delta \alpha)\cdot T(\bm{W}( \bm{x}, \Delta \bm{t})) +
(\beta + \Delta \beta) \Big]^2,
\end{aligned}
\end{equation}
where $T(\bm{x})$ and $I(\bm{x})$ stand for the template image and the current image intensities at position $\bm{x}$, respectively.
The vector $\bm{x}$ iterates through one feature's rectangular neighborhood ($\mathcal{N}$).
We can organize the coefficients of the incremental terms into vector form as
\begin{gather}
\boldsymbol{U}(\boldsymbol{x}) =
\begin{bmatrix}
(1+\alpha)\frac{\partial T(\boldsymbol{x})}{\partial x} \frac{\partial \boldsymbol{W}(\boldsymbol{x}, \boldsymbol{t})}{\partial t_x} \\
(1+\alpha)\frac{\partial T(\boldsymbol{x})}{\partial y} \frac{\partial \boldsymbol{W}(\boldsymbol{x}, \boldsymbol{t})}{\partial t_y} \\
T(\boldsymbol{x}) \\
1
\end{bmatrix},
\end{gather}
and the minimization problem can be rewritten as
\begin{equation}
\min \sum_{\bm{x} \in N}
\Big[ 
(1+\alpha)T(\bm{x}) + \beta - I(\bm{W}(\bm{x},\bm{t})) +
\bm{U}^\intercal(\bm{x}) \Delta \bm{q}
\Big]^2 .
\label{eq:lk_final}
\end{equation}
After computing the derivative of (\ref{eq:lk_final}) and setting it to zero, the solution to $\Delta \bm{q}$ is found using the Hessian $\bm{H}$ by
\begin{gather}
\Delta \bm{q} =
\bm{H}^{-1} \sum_{\bm{x} \in N} \bm{U}^\intercal (\bm{x})\left[  I(\bm{W}(\bm{x},\bm{t})) - (1+\alpha)T(\bm{x}) - \beta \right] \nonumber \\
\bm{H}(\bm{x}) = \sum_{\bm{x} \in N} \bm{U}^\intercal(\bm{x})\bm{U}(\bm{x}). \label{eq:lk_final_solution}
\end{gather}
%

For this algorithm, there are two GPU problems that need to be addressed: memory coalescing and warp divergence.
The problems are approached from the viewpoint of VIO algorithms, where one generally does not track a high number of features (only 50-200), and these sparse features are also scattered throughout the image, which means that they are scattered in memory.

This algorithm minimizes the photometric error in a rectangular neighborhood around each feature on multiple pyramid levels. Consequently, if threads within a warp processed different features, the memory accesses would be uncoalesced, and given some feature tracks do not converge or the number of iterations on the same level differs, some threads within a warp would be idle.
To address both of these concerns, one entire warp is launched for processing one feature.
We also opted for rectangular patch sizes that can be collaboratively processed by warps: on higher resolutions 16x16, on lower resolutions 8x8 pixels.
It solves warp divergence since threads within the warp perform the same number of iterations and they iterate until the same pyramid level.
The memory requests from the warp are also split into fewer memory transactions, as adjacent threads are processing consecutive pixels or consecutive lines.

\begin{figure}[t]
\centering
\includegraphics[width=0.65\linewidth, keepaspectratio]{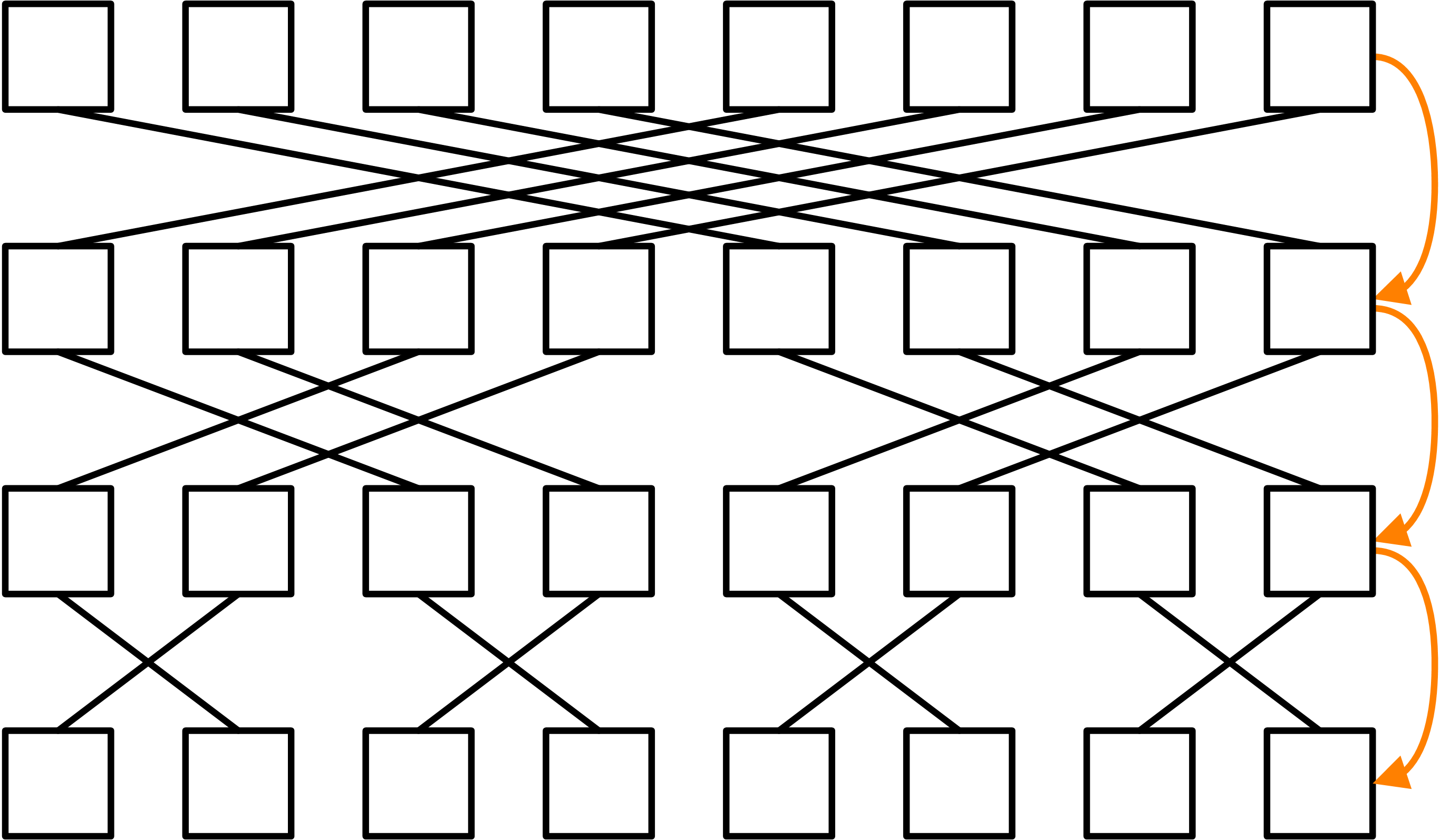}
\caption{Warp-level communication pattern during each minimization iteration, after which each thread has the sum of all individual thread results.}
\label{fig:butterfly}
\end{figure}

Warp-level primitives are exploited in every iteration of the minimization, as we need to perform a patch-wide reduction: in (\ref{eq:lk_final_solution}) we need to sum a four element vector $\bm{U}(\bm{x})^T r(\bm{x},\bm{t})$ for every pixel within the patch.
One thread processes multiple pixels within the patch, hence each thread reduces multiple elements into its registers prior to any communication.
Once the entire warp finishes a patch, the threads need to share their local results with the rest of the warp to calculate $\Delta \boldsymbol{q}$.
This reduction is performed using the butterfly communication pattern shown in Figure \ref{fig:butterfly}.

The novelty of our approach lies in the thread-to-feature assignment.
Approaches presented in \cite{5457608, 5354093} are using a one-to-one assignment, which implies that only large feature-counts can utilize a large number of threads.
This adversely affects latency hiding on GPUs with smaller feature counts, which is generally applicable to VIO.
Our method speeds up the algorithm by having warps that collaboratively solve feature patches, where each thread's workload is reduced, while the used communication medium is the fastest possible.

\section{Evaluation}
We evaluate in four parts: non-maxima suppression, standalone feature detection, feature tracking (all on the EuRoC Machine Hall 01 sequence \cite{Burri25012016}, including 3,682 image frames), and applicability within a VIO pipeline.
The full VIO pipeline is implemented with the bundle-adjustment from \cite{Liu2018cvpr} and tested on the Machine Hall EuRoC dataset sequences \cite{Burri25012016}.

\subsection{Hardware}
We performed our experiments on an NVIDIA Jetson TX2 and a laptop computer with an Intel i7-6700HQ processor and a dedicated NVIDIA 960M graphics card.
The Jetson TX2 was chosen because of its excellent tradeoff between size, weight, and computational capabilities, where we run all experiments with the platform in \textit{max-N} performance mode (all cores and GPU at maximal clock speeds).
The properties of the two platforms are summarized in Table \ref{tab:jetson_tx2_comp}.

\begin{table}[t]
\centering
\caption{Comparison of GPU evaluation hardware}
\label{tab:jetson_tx2_comp}
\vspace{-6pt}
\begin{tabularx}{\linewidth}{l|Y|Y} \hline
\multicolumn{1}{c|}{\textbf{}}        & \textbf{Tegra X2}            & \textbf{960M} \\ \hline \hline
\textbf{GPU Tier}                     & embedded                     & notebook       \\
\textbf{CUDA Capability}              & 6.2                          & 5.0           \\
\textbf{CUDA Cores}                   & 256                          & 640           \\
\textbf{Maximum GPU Clock}            & 1300 MHz                     & 1176 MHz      \\
\textbf{Single-precision Performance} & 665.6 GF/s                   & 1505.28 GF/s  \\
\textbf{Memory Bandwidth}             & 59.7 GB/s                    & 80 GB/s       \\ \hline
\end{tabularx}
\vspace{-12pt}
\end{table}

\subsection{Non-Maxima Suppression}
\begin{table}[b]
\centering
\caption{Comparison of 2D non-maxima suppression kernels on GPU}
\vspace{-5pt}
\label{tab:nms_timings}
\begin{tabular}{l|c|c|c|c} \hline
                       & \multicolumn{2}{c|}{\textbf{Tegra X2}} & \multicolumn{2}{c}{\textbf{960M}} \\ \hline \hline
\textbf{NMS method} & Förstner & Ours & Förstner & Ours \\ \hline
\textbf{n=1 (3x3)} & 294.03 $\mu s$ & \textbf{141.36} $\boldsymbol{\mu s}$ & 96.81 $\mu s$ & \textbf{73.78} $\boldsymbol{\mu s}$ \\ 
\textbf{n=2 (5x5)} & 696.57 $\mu s$ & \textbf{338.03} $\boldsymbol{\mu s}$ & 245.96 $\mu s$ & \textbf{158.93} $\boldsymbol{\mu s}$ \\ 
\textbf{n=3 (7x7)} & 1207 $\mu s$ & \textbf{604.94} $\boldsymbol{\mu s}$ & 441.59 $\mu s$ & \textbf{288.39} $\boldsymbol{\mu s}$ \\
\textbf{n=4 (9x9)} & 1772 $\mu s$ & \textbf{929.82} $\boldsymbol{\mu s}$ & 661.90 $\mu s$ & \textbf{450.08} $\boldsymbol{\mu s}$ \\ \hline
\end{tabular}
\end{table}

The proposed FAST corner response function ($\epsilon=10$, $N=10$) was used as input to our non-maxima suppression, run with a grid granularity of 32$\times$32 $(l=1, w=1, h=32)$ and compared to  Förstner's \cite{forstner1987fast} spiral non-maxima suppression algorithm.
The results are listed in Table~\ref{tab:nms_timings}, showing a 2$\times$ speedup for the embedded Jetson TX2 platform.

\subsection{Feature detector}
\subsubsection{Conformance}
\begin{figure}[t]
\centering
\subfigure[Sum of absolute differences \cite{rosten_2006_machine}\label{pics:conformance_sad}]{\includegraphics[width=0.2375\textwidth, keepaspectratio]{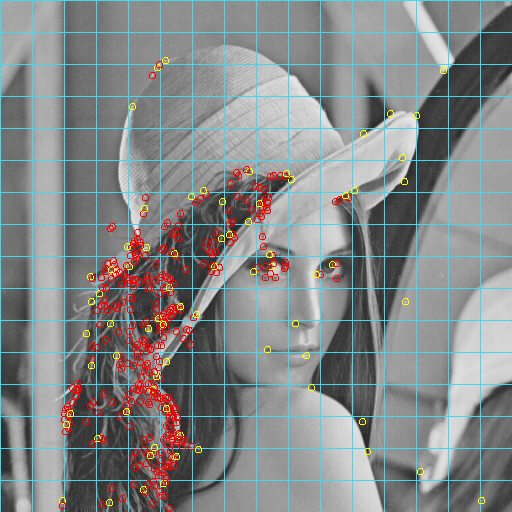}}
\subfigure[Maximum threshold value \cite{rosten_2008_faster}\label{pics:conformance_mtv}]{\includegraphics[width=0.2375\textwidth, keepaspectratio]{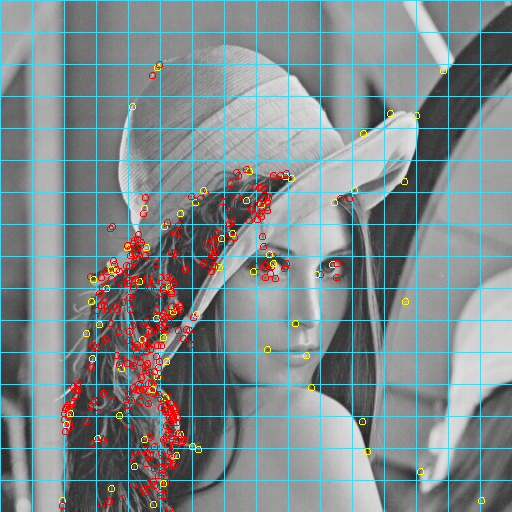}}
\vspace{-6pt}
\caption{Conformance verification with the original FAST detector ({\color{red}$\circ$}) and
our combined FAST/NMS ({\color{yellow}$\circ$} for conforming detections, {\color{blue}$\circ$} for false-positives).
Note that there are no false-positives and that our method returns a well-distributed subset of the original response, therefore rendering additional feature selection unnecessary. Best viewed in digital paper.}
\label{pics:conformance}
\vspace{-10pt}
\end{figure}

We verify our feature detection conformance with the original FAST feature detector, for both suggested score functions:
\ref{pics:conformance_sad} the sum of the absolute difference between the center pixel and the contiguous arc;
\ref{pics:conformance_mtv} the maximum threshold value, for which the point is detected as a corner. 
As our combined non-maxim suppression selects a single maximum within each cell, our output only comprises of a subset of the original implementation's output.
In Figure \ref{pics:conformance} we mark features red~{\color{red}$\circ$} which are output from the original detector, yellow~{\color{yellow}$\circ$} which are output from both implementations, and blue~{\color{blue}$\circ$} false-positives of our detector.
Note that there are no false-positives and that our method returns a well-distributed subset of the original response, rendering additional feature selection unneccessary,

\subsubsection{Cache-hit ratio}
As mentioned in \ref{sec:fast}, we expect higher GPU cache-hit ratios during feature detection with our bit-based CRF lookup table.
This reduces the number of global-memory transactions, resulting in lower kernel execution times.
The cache-hit ratios and the resulting CRF timings are listed in Table \ref{tab:fast_lut_crf}.

\begingroup
\setlength{\tabcolsep}{2pt}
\begin{table}[b]
\centering
\caption{Timing Comparison of different FAST CRF Scores}
\vspace{-5pt}
\label{tab:fast_lut_crf}
\begin{tabular}{ll|c|c|c|c} \hline
\multicolumn{2}{c|}{} & \multicolumn{2}{c|}{\textbf{Tegra X2}} & \multicolumn{2}{c}{\textbf{960M}} \\ \hline \hline
\multicolumn{2}{l|}{\textbf{Lookup table}} & byte-based & bit-based & byte-based & bit-based \\ \hline
\multicolumn{1}{c|}{\multirow{2}{*}{SAD-B}} & \textbf{L1 cache-hit rate} & 83.9 \% & \textbf{89.9 \%} &  68.6 \% & \textbf{77.4 \%} \\
\multicolumn{1}{c|}{} & \textbf{CRF kernel} & 317.3 $\mu s$ & \textbf{298.7} $\boldsymbol{\mu s}$ & 141.3 $\mu s$ & \textbf{135.9} $\boldsymbol{\mu s}$ \\ \hline
\multicolumn{1}{c|}{\multirow{2}{*}{SAD-A}} & \textbf{L1 cache-hit rate} & 83.9 \% &  \textbf{89.8 \%} & 68.5 \% & \textbf{77.3 \%} \\
\multicolumn{1}{c|}{} & \textbf{CRF kernel} & 348.4 $\mu s$ & \textbf{334.9} $\boldsymbol{\mu s}$ & 158.5 $\mu s$ & \textbf{155.1} $\boldsymbol{\mu s}$ \\ \hline
\multicolumn{1}{c|}{\multirow{2}{*}{MT}} & \textbf{L1 cache-hit rate} & 84.0 \% & \textbf{91.8 \%} & 71.9 \% & \textbf{82.7 \%} \\
\multicolumn{1}{c|}{} & \textbf{CRF kernel} & 815.1 $\mu s$ & \textbf{784.3} $\boldsymbol{\mu s}$ & 410.6 $\mu s$ & \textbf{374.8} $\boldsymbol{\mu s}$ \\ \hline
\end{tabular}
\end{table}
\endgroup

\subsubsection{Execution time breakdown}
We split the execution time of pyramidal feature detection ($l=2$) into its constituents:
image copy from host to device memory (\textit{Upload}),
creation of an image pyramid where each subsequent layer halves the resolution of the previous one (\textit{Pyramid}),
corner response function evaluation (\textit{CRF}),
non-maxima suppression with cell-maximum selection (\textit{NMS}),
and feature grid copy from device memory to host memory (\textit{Download}).

\begin{figure}[t]
\centering
\includegraphics[width=0.47\textwidth, trim=0 8 0 0, clip, keepaspectratio]{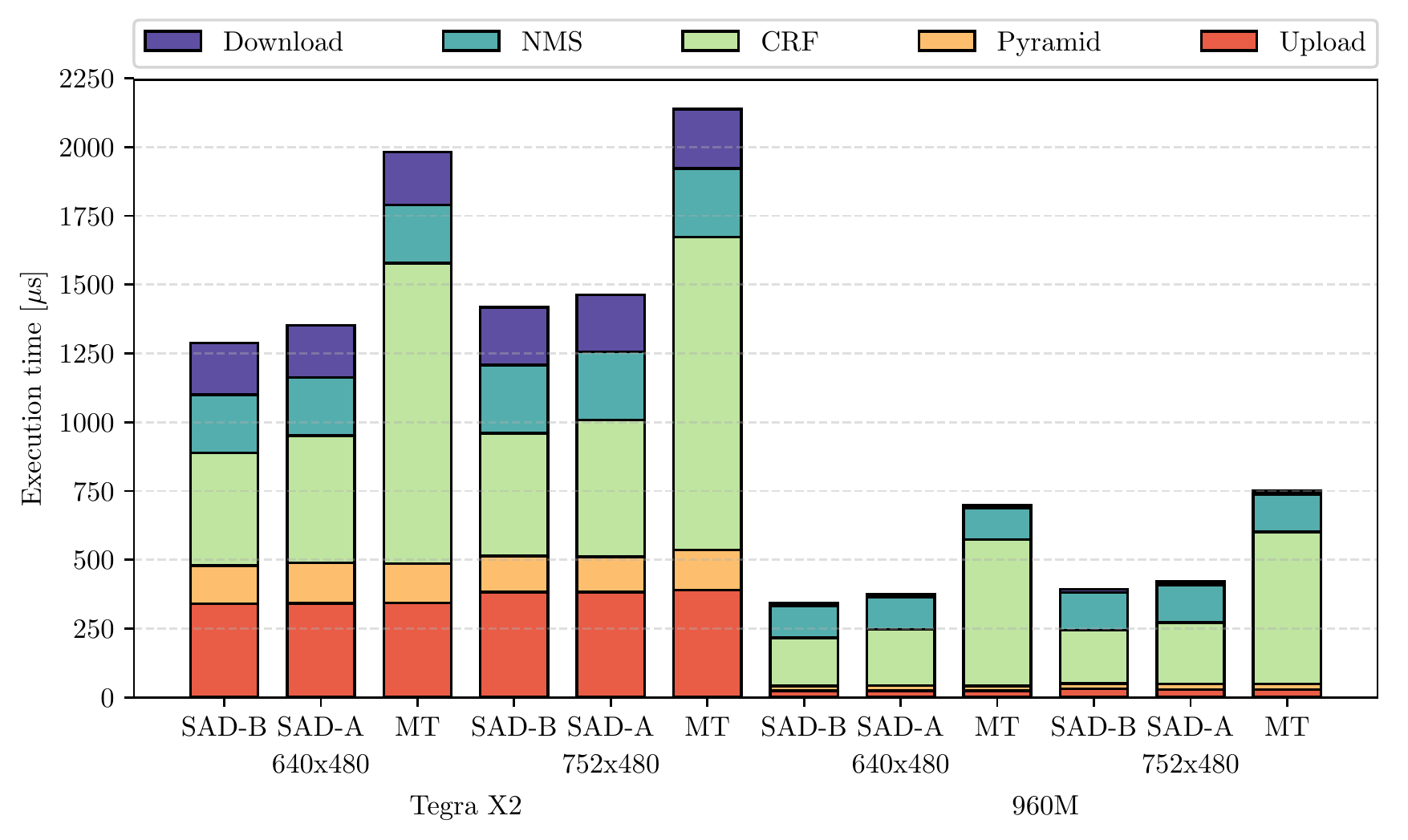}
\vspace{-8pt}
\caption{Feature detector execution time breakdown into image upload, pyramid creation, CRF, NMS, and feature download. Best viewed in color.}
\vspace{-8pt}
\end{figure}

\subsubsection{Execution time comparison}
We compare our feature detection with other publicly available FAST implementations, summarized in Table \ref{tab:feat_det_comparison}.
As the publicly available detectors only support single-scale, we performed these experiments on the original image resolution.
Note that our method not only performs feature extraction but simultaneously applies cell-wise non-maxima suppression, still achieving superior execution times.
As KFAST uses x86-specific instruction set extensions, while ArrayFire OpenCL was incompatible with our available packages, these could not be run on the Jetson TX2.
The GPU timings include the image upload and feature list download times as well.

\begingroup
\setlength{\tabcolsep}{3pt}
\begin{table}[b]
\vspace{-5pt}
\centering
\caption{FAST feature detector average execution time comparison}
\vspace{-5pt}
\begin{tabularx}{\linewidth}{l|Y|Y}
\hline
& \textbf{Tegra X2} & \textbf{960M} \\
\hline
\hline
\multicolumn{3}{c}{\textbf{Others} (feature detection and NMS)} \\ \hline
\textbf{OpenCV CPU} with MT & 4.78 ms & 2.23 ms \\
\textbf{OpenCV CUDA}  with MT & 2.73 ms & 1.09 ms \\
\textbf{ArrayFire CPU}  with SAD-A & 60.83 ms & 36.42 ms \\
\textbf{ArrayFire CUDA} with SAD-A & 1.47 ms & 0.51 ms \\
\textbf{ArrayFire OpenCL} with SAD-A & - & 0.91 ms \\
\textbf{KFAST CPU} with MT & - & 0.63 ms \\
\hline
\multicolumn{3}{c}{\textbf{Ours} (feature detection, NMS, and NMS-C)} \\ \hline
\textbf{Ours} with SAD-B & 1.11 ms & 0.28 ms \\
\textbf{Ours} with SAD-A & 1.14 ms & 0.30 ms \\
\textbf{Ours} with MT & 1.59 ms & 0.52 ms \\
\hline
\end{tabularx}
\label{tab:feat_det_comparison}
\end{table}
\endgroup

\subsection{Feature tracker}
\begin{figure}[t]
\centering
\subfigure[Performance on 960M]{\includegraphics[width=0.43\textwidth,
trim=0 8 0 0, clip, keepaspectratio]{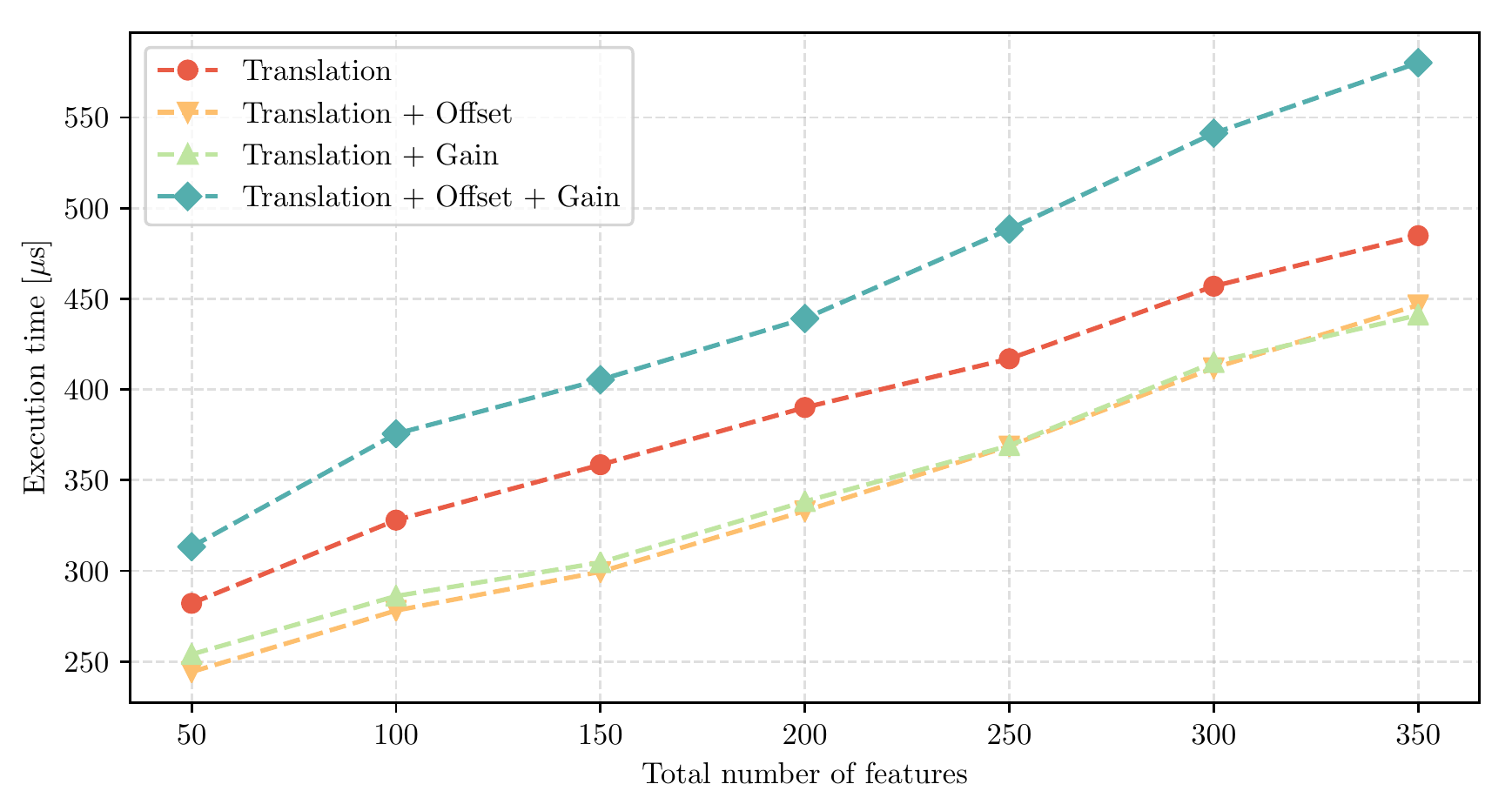}}
\subfigure[Performance on Tegra X2]{\includegraphics[width=0.43\textwidth, trim=0 8 0 0, clip, keepaspectratio]{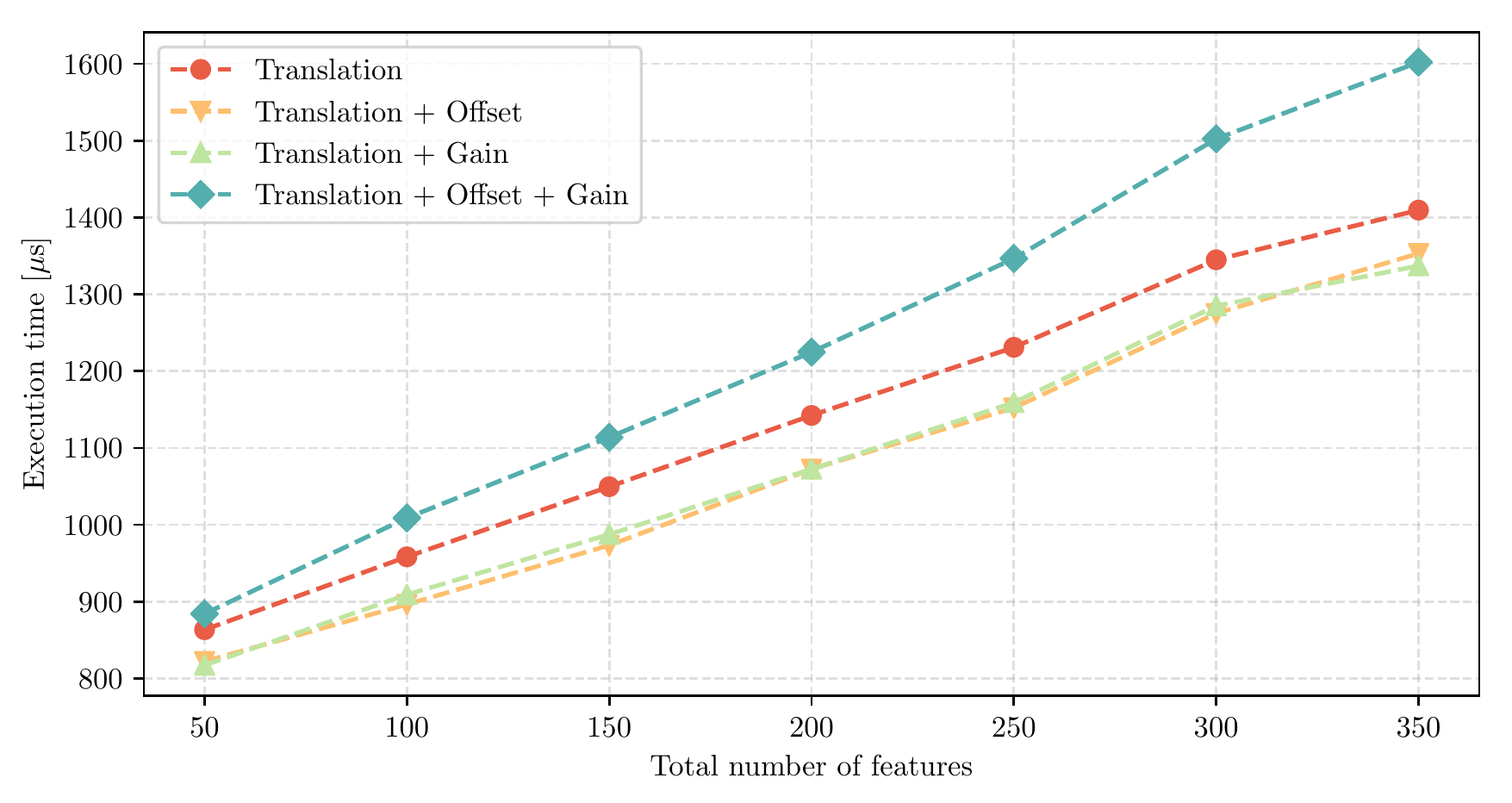}}
\vspace{-8pt}
\caption{Feature tracker performance comparison over a varying number of tracked features, with translation, illumination gain and offset estimation, and subsets of those. Best viewed in color.}
\label{pics:tracker_timing}
\vspace{-8pt}
\end{figure}

We timed our feature tracker implementation by varying the total number of simultaneous feature tracks on both testing platforms, depicted in Figure \ref{pics:tracker_timing}.
We utilized our FAST with score (ii) as feature detector, and triggered feature re-detection whenever the feature track count falls below 30\% of the actual target.
The evaluations include the full affine intensity and translation estimation, with a total of 4 estimated parameters.
We also show the translation-only estimation, translation-gain, as well as translation-offset estimation.
Note that translation-only estimation is less efficient since it needs more iterations to converge, as visible in Figure. \ref{pics:tracker_timing}.
Lastly, Table \ref{tab:feature_tracker} shows a comparison against OpenCV's CPU and GPU implementation, which we outperform by a factor of 2$\times$ and more.

\begingroup
\setlength{\tabcolsep}{3pt}
\begin{table}[b]
\centering
\vspace{-6pt}
\caption{Feature tracker average execution time comparison, tracking 100 features, re-detection at 30 features}
\vspace{-5pt}
\begin{tabularx}{\linewidth}{l|Y|Y}
\hline
& \textbf{Tegra X2} & \textbf{i7-6700HQ+960M} \\
\hline
\hline
\textbf{OpenCV CPU} & 1.88 ms & 1.38 ms \\
\textbf{OpenCV CUDA}  & 3.96 ms & 0.74 ms \\
\hline
\textbf{Ours} trans. only & 0.96 ms & 0.33 ms \\ 
\textbf{Ours} trans. \& offset & 0.90 ms & 0.28 ms \\ 
\textbf{Ours} trans. \& gain & 0.91 ms & 0.29 ms \\
\textbf{Ours} trans. \& gain \& offset & 1.01 ms & 0.38 ms \\ 
\hline
\end{tabularx}
\label{tab:feature_tracker}
\end{table}
\endgroup

\subsection{Visual odometry}
Lastly, we evaluate the performance of our frontend in combination with a VIO bundle-adjustment backend.
We chose ICE-BA \cite{Liu2018cvpr} as backend, since it is accurate, efficient and also achieves extremely fast execution times.
We implement two test cases: first, we run the ICE-BA backend with their proposed frontend as a baseline and then compare it against our frontend with their backend.
Both frontends employ FAST features (our implementation vs. theirs), a Lucas-Kanade feature tracker with 70 tracked features.
We tuned the original ICE-BA configuration \cite{Liu2018cvpr} and reduced the local bundle adjustment (LBA) window size to 15 frames for both cases, while we kept other parameters unchanged.
We summarize our findings in Table. \ref{tab:vio_results}, which shows an average speedup factor of $2.25\times$ of the combined front- and back-end on both platforms, with an average accuracy loss of $0.47\%$.
Our tracking performance on MH\textunderscore04 and MH\textunderscore05 is affected by the faster motions and dark scenes.
However, according to \cite{Delmerico2018},
most pipelines suffer from largely increased tracking error on these two sequences.
This is due to the relatively dark appearance combined with fast motions of some scenes in these sequences, introducing higher noise in feature tracking.
The VIO pipeline combining our GPU-accelerated frontend and the CPU targetted ICE-BA backend allows us to achieve a throughput of $\sim$200fps on multiple datasets on the embedded Jetson TX2 platform.

\begin{table*}[t]
\centering
\caption{Results of our frontend combined with a VIO backend \cite{Liu2018cvpr} achieving $\sim$200fps throughput on the EuRoC dataset \cite{Burri25012016}}
\vspace{-5pt}
\label{tab:vio_results}
\begin{tabular}{l|l|ccccc|ccccc}
\hline
 & & \multicolumn{5}{c|}{\textbf{Average execution time}} & \multicolumn{5}{c}{\textbf{Relative translation error (RMSE)}} \\ \hline
 & & MH\_01   & MH\_02   & MH\_03   & MH\_04   & MH\_05   & MH\_01      & MH\_02     & MH\_03      & MH\_04     & MH\_05     \\ \hline
\multicolumn{1}{l|}{\multirow{2}{*}{\textbf{Tegra X2}}} & \textbf{Original} & 11.64 ms & 12.90 ms & 12.91 ms & 12.90 ms & 12.85 ms & 1.08 \% & 0.71 \% & 0.40 \% & 0.81 \% & 0.50 \% \\
 & \textbf{Ours} & 4.67 ms  & 4.93 ms  & 6.41 ms  & 6.10 ms  & 5.87 ms  & 0.86 \% & 0.90 \% & 1.40 \% & 1.85 \% & 1.19 \% \\ \hline
\multicolumn{1}{l|}{\multirow{2}{*}{\textbf{i7-6700HQ+960M}}} & \textbf{Original} & 4.11 ms & 4.28 ms & 4.77 ms & 4.63 ms & 4.64 ms & 0.68 \% & 0.71 \% & 0.53 \% & 0.83 \% & 1.36 \% \\
 & \textbf{Ours} & 1.56 ms  & 1.74 ms  & 2.54 ms  & 2.51 ms  & 2.06 ms  & 1.00 \% & 0.55 \% & 0.74 \% & 2.14 \% & 1.68 \% \\ \hline
\end{tabular}
\vspace{-5pt}
\end{table*}

\section{Conclusion}
This work introduces a novel non-maxima suppression exploiting low-level GPU-specific instruction primitives, complemented by a GPU-accelerated FAST feature detector implementing multiple corner response functions.
Our approach is unique in the way it combines feature detection and non-maxima suppression, not only guaranteeing uniform feature distribution over an image but also consistently outperforming all other available implementations in terms of execution speed. The speed improvement for the embedded computer is more pronounced, as there's a higher performance penalty for memory interactions than in the case of the laptop GPU.
We verified the conformity with the original FAST detector, analyzed the execution timings on two different platforms considering corner response functions, non-maxima suppression, and feature tracking on multiple numbers of features.
As opposed to others, our feature tracker utilizes a feature-to-warp assignment, which speeds up tracking operations in typical VIO scenarios.
Finally, we demonstrate superior speed in combining our frontend with a VIO bundle-adjustment backend, achieving a metric state estimation throughput of $\sim$200 frames per second with high accuracy on an embedded Jetson TX2 platform, providing a real-time, heterogeneous VIO alternative.

{\tiny
\balance
\bibliographystyle{unsrt}
\bibliography{references}

\begin{thebibliography}{10}

\bibitem{Harris88acombined}
C.~Harris and M.~Stephens.
\newblock A combined corner and edge detector.
\newblock In {\em In Proc. of Fourth Alvey Vision Conference}, 1988.

\bibitem{Shi:1993:GFT:866676}
Jianbo S. and Carlo T.
\newblock Good features to track.
\newblock {\em {IEEE} Conf. Comput. Vis. Pattern Recog. (CVPR)}, 1994.

\bibitem{rosten_2006_machine}
E.~Rosten and T.~Drummond.
\newblock Machine learning for high-speed corner detection.
\newblock In {\em Eur. Conf. Comput. Vis. (ECCV)}, 2006.

\bibitem{Rublee:2011:OEA:2355573.2356268}
E.~Rublee, V.~Rabaud, K.~Konolige, and G.~Bradski.
\newblock Orb: An efficient alternative to sift or surf.
\newblock In {\em Int. Conf. Comput. Vis. (ICCV)}, 2011.

\bibitem{Lowe:2004:DIF:993451.996342}
David~G. Lowe.
\newblock Distinctive image features from scale-invariant keypoints.
\newblock {\em Int. J. Comput. Vis.}, 2004.

\bibitem{Bay:2008:SRF:1370312.1370556}
H.~Bay, A.~Ess, T.~Tuytelaars, and L.~Van~Gool.
\newblock Speeded-up robust features (surf).
\newblock {\em Comput. Vis. Image. Und.}, 2008.

\bibitem{rosten_2008_faster}
E.~Rosten, R.~Porter, and T.~Drummond.
\newblock Faster and better: A machine learning approach to corner detection.
\newblock {\em {IEEE} Trans. Pattern Anal. Mach. Intell.}, 2010.

\bibitem{Komrad2006github}
K.~Omar.
\newblock Kfast: vectorized x86 cpu implementation of the fast feature
  detector, 2006.

\bibitem{opencv_library}
G.~Bradski.
\newblock {The OpenCV Library}.
\newblock {\em Dr. Dobb's Journal of Software Tools}, 2000.

\bibitem{Yalamanchili2015}
P.~Yalamanchili, U.~Arshad, Z.~Mohammed, P.~Garigipati, P.~Entschev,
  B.~Kloppenborg, J.~Malcolm, and J.~Melonakos.
\newblock {ArrayFire - A high performance software library for parallel
  computing with an easy-to-use API}, 2015.

\bibitem{1699659}
A.~{Neubeck} and L.~{Van Gool}.
\newblock Efficient non-maximum suppression.
\newblock In {\em {IEEE} Int. Conf. Pattern Recog. (ICPR)}, 2006.

\bibitem{Pham10}
Tuan~Q. Pham.
\newblock Non-maximum suppression using fewer than two comparisons per pixel.
\newblock In {\em Advanced Concepts for Intelligent Vision Systems {ACIVS}},
  2010.

\bibitem{7471831}
D.~{Oro}, C.~{Fernández}, X.~{Martorell}, and J.~{Hernando}.
\newblock Work-efficient parallel non-maximum suppression for embedded gpu
  architectures.
\newblock In {\em IEEE International Conference on Acoustics, Speech and Signal
  Processing (ICASSP)}, 2016.

\bibitem{forstner1987fast}
W.~F\"orstner and E.~G\"ulch.
\newblock A fast operator for detection and precise location of distinct point,
  corners and centres of circular features.
\newblock In {\em Proceedings of the ISPRS Conference on Fast Processing of
  Photogrammetric Data}, 1987.

\bibitem{4209642}
A.~I. {Mourikis} and S.~I. {Roumeliotis}.
\newblock A multi-state constraint kalman filter for vision-aided inertial
  navigation.
\newblock In {\em {IEEE} Int. Conf. Robot. Autom. (ICRA)}, 2007.

\bibitem{Leutenegger:2015:KVO:2744155.2744163}
S.~Leutenegger, S.~Lynen, M.~Bosse, R.~Siegwart, and P.~Furgale.
\newblock Keyframe-based visual-inertial odometry using nonlinear optimization.
\newblock {\em Int. J. Robot. Research}, 2015.

\bibitem{journals/corr/Mur-ArtalMT15}
Ra\'{u}l Mur-Artal, Jos\'e M.~M. Montiel, and Juan~D. Tard{\'o}s.
\newblock {ORB-SLAM}: a versatile and accurate monocular {SLAM} system.
\newblock {\em {IEEE} Trans. Robot.}, 2015.

\bibitem{7353389}
M.~Bloesch, S.~Omari, M.~Hutter, and R.~Siegwart.
\newblock Robust visual inertial odometry using a direct {EKF}-based approach.
\newblock In {\em IEEE/RSJ Int. Conf. Intell. Robot. Syst. (IROS)}, 2015.

\bibitem{Lucas81ijcai}
B.~Lucas and T.~Kanade.
\newblock An iterative image registration technique with an application to
  stereo vision.
\newblock In {\em Int. Joint Conf. Artificial Intell. (IJCAI)}, 1981.

\bibitem{Baker-2002-8493}
S.~Baker and I.~Matthews.
\newblock Lucas-kanade 20 years on: A unifying framework: Part 1.
\newblock {\em Int. J. Comput. Vis.}, 2002.

\bibitem{Baker-2003-8809}
S.~Baker, R.~Gross, and I.~Matthews.
\newblock Lucas-kanade 20 years on: A unifying framework: Part 3.
\newblock {\em Int. J. Comput. Vis.}, 2003.

\bibitem{6906584}
C.~{Forster}, M.~{Pizzoli}, and D.~{Scaramuzza}.
\newblock Svo: Fast semi-direct monocular visual odometry.
\newblock In {\em {IEEE} Int. Conf. Robot. Autom. (ICRA)}, 2014.

\bibitem{Forster:2017:SSV:3083770.3083863}
C.~Forster, Z.~Zhang, M.~Gassner, M.~Werlberger, and D.~Scaramuzza.
\newblock {SVO:} semidirect visual odometry for monocular and multicamera
  systems.
\newblock {\em {IEEE} Trans. Robot.}, 2017.

\bibitem{4563089}
C.~{Zach}, D.~{Gallup}, and J.~{Frahm}.
\newblock Fast gain-adaptive klt tracking on the gpu.
\newblock In {\em {IEEE} Conf. Comput. Vis. Pattern Recog. Workshops (CVPRW)},
  2008.

\bibitem{5457608}
J.S. Kim, M.~Hwangbo, and T.~Kanade.
\newblock Realtime affine-photometric klt feature tracker on gpu in cuda
  framework.
\newblock In {\em Int. Conf. Comput. Vis. Workshops (ICCVW)}, 2009.

\bibitem{5354093}
M.~{Hwangbo}, J.~{Kim}, and T.~{Kanade}.
\newblock Inertial-aided klt feature tracking for a moving camera.
\newblock In {\em IEEE/RSJ Int. Conf. Intell. Robot. Syst. (IROS)}, 2009.

\bibitem{Burri25012016}
M.~Burri, J.~Nikolic, P.~Gohl, T.~Schneider, J.~Rehder, S.~Omari, M.W.
  Achtelik, and R.~Siegwart.
\newblock The {EuRoC} micro aerial vehicle datasets.
\newblock {\em Int. J. Robot. Research}, 2015.

\bibitem{cudacprogrammingguide}
NVIDIA Corporation.
\newblock {\em CUDA C++ Programming Guide}, 2019.

\bibitem{1315094}
D.~Nister, O.~Naroditsky, and J.~Bergen.
\newblock Visual odometry.
\newblock In {\em {IEEE} Conf. Comput. Vis. Pattern Recog. (CVPR)}, 2004.

\bibitem{Scaramuzza:2009:RMV:1703435.1703515}
D.~Scaramuzza, F.~Fraundorfer, and R.~Siegwart.
\newblock Real-time monocular visual odometry for on-road vehicles with 1-point
  ransac.
\newblock In {\em {IEEE} Int. Conf. Robot. Autom. (ICRA)}, 2009.

\bibitem{6153423}
F.~{Fraundorfer} and D.~{Scaramuzza}.
\newblock Visual odometry : Part ii: Matching, robustness, optimization, and
  applications.
\newblock {\em {IEEE} Robot. Autom. Mag.}, 2012.

\bibitem{devblogfasterparallelreduction}
NVIDIA Corporation.
\newblock {\em Developer Blog - Faster Parallel Reductions on Kepler}, 2014.
\newblock \url{https://devblogs.nvidia.com/faster-parallel-reductions-kepler/}.

\bibitem{Liu2018cvpr}
H.~{Liu}, M.~{Chen}, G.~{Zhang}, H.~{Bao}, and Y.~{Bao}.
\newblock Ice-ba: Incremental, consistent and efficient bundle adjustment for
  visual-inertial slam.
\newblock In {\em {IEEE} Conf. Comput. Vis. Pattern Recog. (CVPR)}, 2018.

\bibitem{Delmerico2018}
Jeffrey Delmerico and Davide Scaramuzza.
\newblock A benchmark comparison of monocular visual-inertial odometry
  algorithms for flying robots.
\newblock In {\em {IEEE} Int. Conf. Robot. Autom. (ICRA)}, 2018.

\end{thebibliography}
}
\end{document}